\documentclass{article}

\usepackage[utf8]{inputenc}
\usepackage[margin=1in]{geometry}
\usepackage[titletoc,title]{appendix}
\usepackage{booktabs}
\usepackage{caption}
\usepackage{subcaption}
\usepackage{fancyvrb}

\RecustomVerbatimCommand{\VerbatimInput}{VerbatimInput}%
{fontsize=\footnotesize,
 frame=lines,  
 framesep=2em, 
 rulecolor=\color{Gray},
 label=\fbox{\color{Black}data.txt},
 labelposition=topline,
 commandchars=\|\(\), 
 commentchar=*        
}

\usepackage{amsmath,amsfonts,amssymb,mathtools}

\usepackage{graphicx,float}
\usepackage[affil-it]{authblk}

\usepackage[ruled,vlined]{algorithm2e}
\usepackage{algorithmic}
\usepackage[utf8]{inputenc}

\usepackage{biblatex}
\title{Use of social media and Natural Language Processing (NLP) in natural hazard research}
\author{José Augusto Proença Maia Devienne}
\affil{Big Data to Earth Scientists}
\date{January, 2021}

\begin{document}

\maketitle

\section{Introduction}

\textit{Twitter} is a microblogging service for sending short, public text messages (\textit{tweets}) that has recently received more attention in scientific community. In the works of \textit{Sasaki et al. (2010)} and \textit{Earle et al. (2011)} the authors explored the real-time interaction on Twitter for detecting natural hazards (e.g., earthquakes, typhoons) based on the user's tweets on twitter. An inherent challenge for such an application is the natural language processing (NLP), which basically consists in converting the words in numbers (vectors and tensors) in order to (mathematically/ computationally) make predictions and classifications. Recently advanced computational tools have been made available for dealing with text computationally. In this report were implemented a NLP machine learning with TensorFlow, an end-to-end open source platform for machine learning,  to process and classify events based on files containing only text.




\section{Web scraping with Selenium}

The first part of this report focused on to automate the process of getting multiple files containing only texts. The data set was obtained from ResearchGate, a social media suited for researchers in which different type of content (poster, paper, projects, questions about specific topics, etc) may be posted and accessed by the other users. 

In a usual search (i.e., a search using a browser such as Chrome, Firefox, etc), a ResearchGate's users should, first of all, log in his/her personal account by inserting his/her e-mail and password. Once logged in, the user has the possibility of looking for different words in different contexts (papers, presentation, questions, etc). Being the ResearchGate a dynamic web page, as the user scroll down the web page, more content is automatically uploaded. The content of each item in the search page can be accessed by simply clicking on the post's name. This is the traditional route each user, in principle, should follow in order to get access to the content of a publication uploaded in ResearchGate. Nevertheless, this process may be tedious and laborious if the interest is to access a large amount of posts for a giving word of interest.

\begin{figure}[h!]
    \centering
    \includegraphics[width=0.7\textwidth]{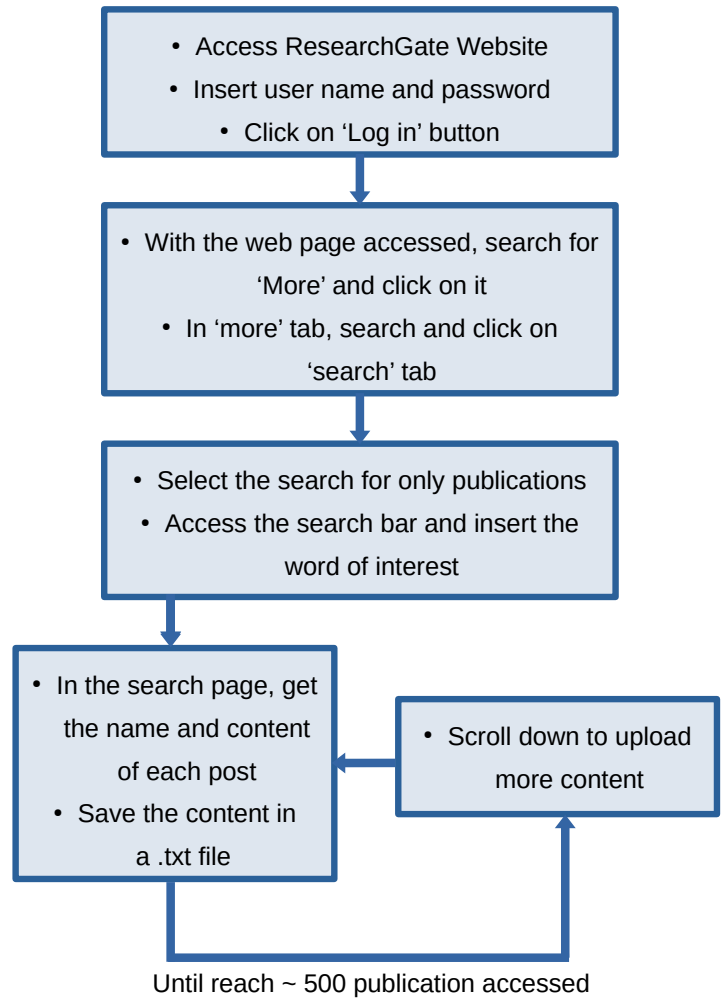}
    \caption{Scheme adopted to upload the content of multiple post on ResearchGate.}
    \label{fig:organogram}
\end{figure}

\begin{figure}[h!]
    \centering
    \includegraphics[width=0.9\textwidth]{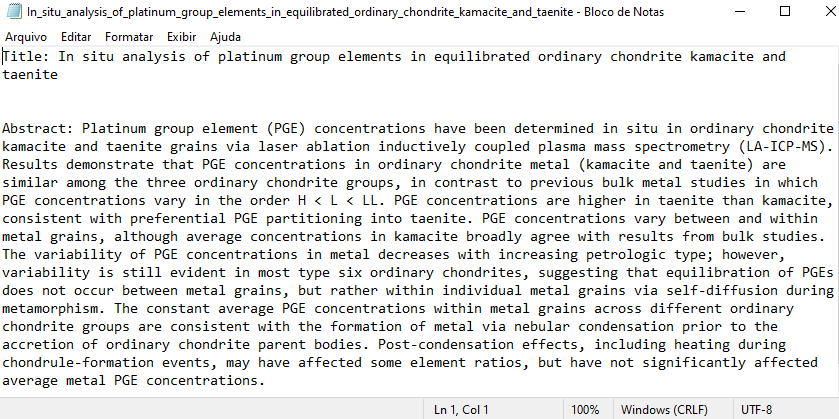}
    \caption{Example of an exported \textit{.txt} file with the content of a publication found in ResearchGate when looking for the word 'taenite' in the search bar.}
    \label{fig:txt}
\end{figure}

In order to automate this process we relied on Selenium. Selenium is a portable framework for testing web applications. It provides a playback tool for authoring functional tests without the need to learn a test scripting language. Selenium can be used with Python as a usual package and, when implemented in Python scripts, allows the access of multiple content/posts of a web page at once. In this process scripts act in the background of an usual browser (in this case, Google Chrome). The list of commands in the script includes the logging in (i.e., inserting the user name and password), searching for the 'search' tab, specifying the search to be done only in the publications (i.e., content uploaded in the category of 'Publication'), inputting the word to be searched (in this case, 'taenite') and submit the search, collecting the content of each post uploaded in the search page, scrolling down for uploading new content and getting the content of the newly uploaded content. This process is continued until $\sim$ 500 publications were uploaded (see figure \ref{fig:organogram} and the file \textit{web\_scraping.py} attached) and its content properly saved and exported as a \textit{.txt} file. An example of the exported files is presented in figure \ref{fig:txt}.

\subsection{Word2Vec}

The second part of this report has focused on the Word2Vec implementation. Word2Vec is a family of model architectures and optimizations that can be used to learn word embedding from large datasets. There are different methods for learning representations of words. In this report the continuous skip-gram model was implemented, which is suitable for predicting words within a certain range before and after the current word in the same sentence. In other words, the skip-gram model is able to predict the contexts based on single words, which means that a word can have multiple meanings (i.e., have different contexts). 

\begin{figure}[h!]
    \centering
    \includegraphics[width = 0.7\textwidth]{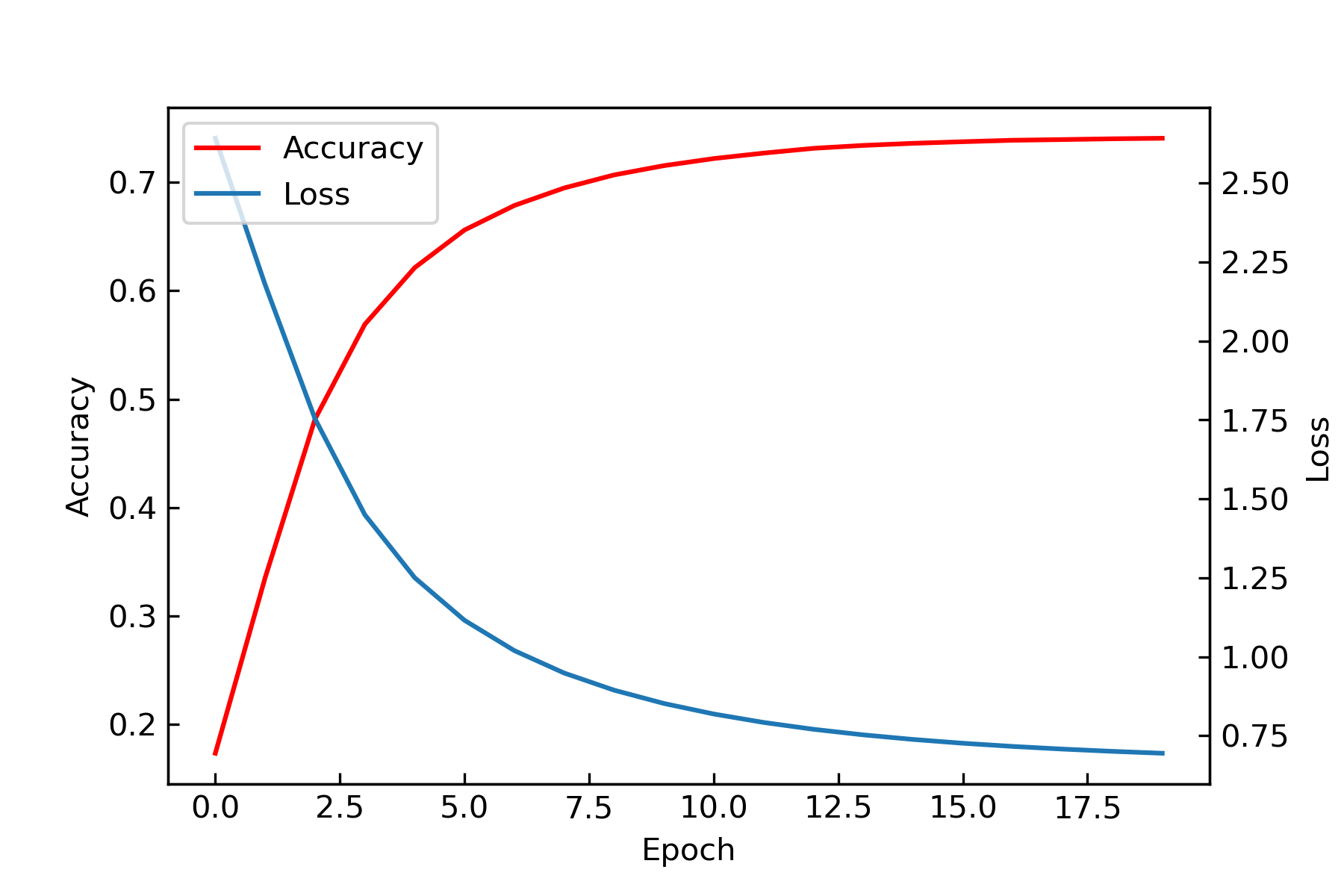}
    \caption{Loss function and model accuracy as a function of the epochs. }
    \label{fig:loss_acc}
\end{figure}

The strategy used to get the content of multiple files from the \textit{corpus} was to concatenate them into a single document (see the attached \textit{word2vec.py}). Next, a standardization functions was used to preprocess the data set (i.e., put all words in lowercase and remove the punctuation). Once the standardization is performed, the text can be properly converted into vectors (text vectorization). With posses of the vectorial representation of the data set, the machine learning model can be constructed and tested. The figure \ref{fig:loss_acc} show the behaviour of the loss function (in this report the loss function utilized was the categorical cross entropy) and the accuracy of the model as a function of the epochs. The accuracy reached $\sim $ 78\%, which represents a good estimate for the context estimation (i.e., nearest neighbors estimation).

\begin{figure}[h!]
    \centering
    \includegraphics[width = 0.9\textwidth, height = 0.57\textwidth]{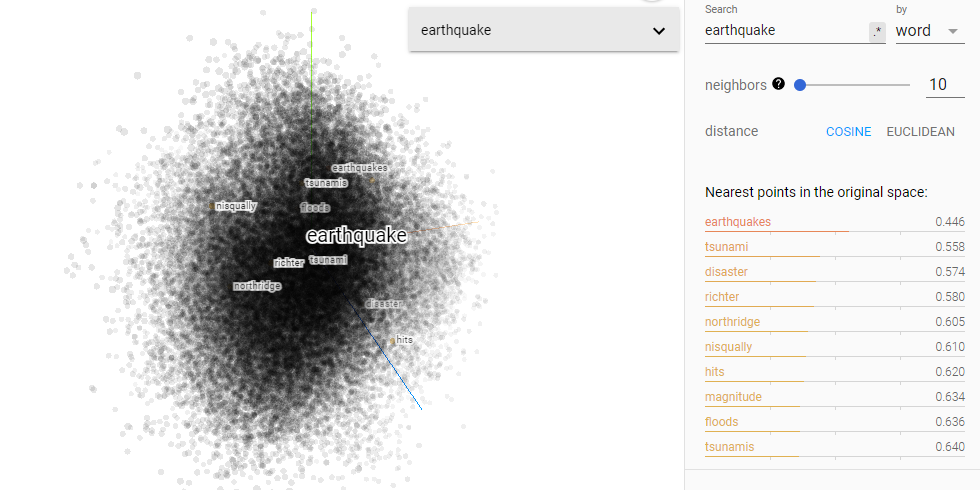}
    \caption{Near neighbors of the word 'earthquake' calculated using cosine distance considering all (vectorized) words.}
    \label{fig:NN}
\end{figure}

\begin{figure}[h!]
    \centering
    \includegraphics[width = 0.9\textwidth, height = 0.6\textwidth]{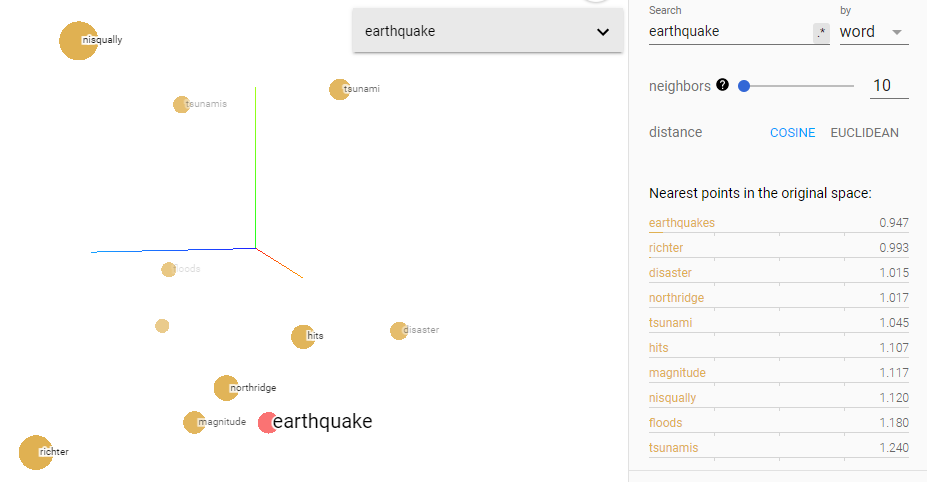}
    \caption{Near neighbors of the word 'earthquake' considering the first ten neighbors exclusively.}
    \label{fig:NN_iso}
\end{figure}

We can assess the accuracy of this methodology with Tensorboard. In figure \ref{fig:NN} and \ref{fig:NN_iso} the word 'earthquake' was selected to have its nearest neighbor estimated. Both figures show the first ten neighbors of 'earthquake', obtained by implementing a principal component analysis (which is possible to be done at Tensorboard as well). In both figues the nearest neighbors have been calculated using cosine distance. It's worthy to mention that the model could correctly find other word with similar meaning at the vicinity of 'earthquake'. Considering that for this step publicatons from both categories ('tsunamis' and 'earthquakes') were analysed together, in both figures is possible to see that both 'tsunamis' and 'earthquakes' have been included in the same 'category' (i.e., both are in the vicinity of 'disaster').

The next part of the embedding implementation was a classification of events (i.e., context) based on words. For this, 500  text files of each category ('tsunamis' and 'earthquakes') were analysed separately. 100 files of each category were selected for training the model, and the rest of the 400 files separated for validation. In figures 8 to 13 the classification was implemented by different values of sequence size, vocabulary size, embedding dimensions and batch size. For all the simulations a fixed value of epochs (15) was adopted. 

\begin{figure}[h!]
  \centering
  \begin{minipage}[b]{0.45\textwidth}
    \includegraphics[width=\textwidth]{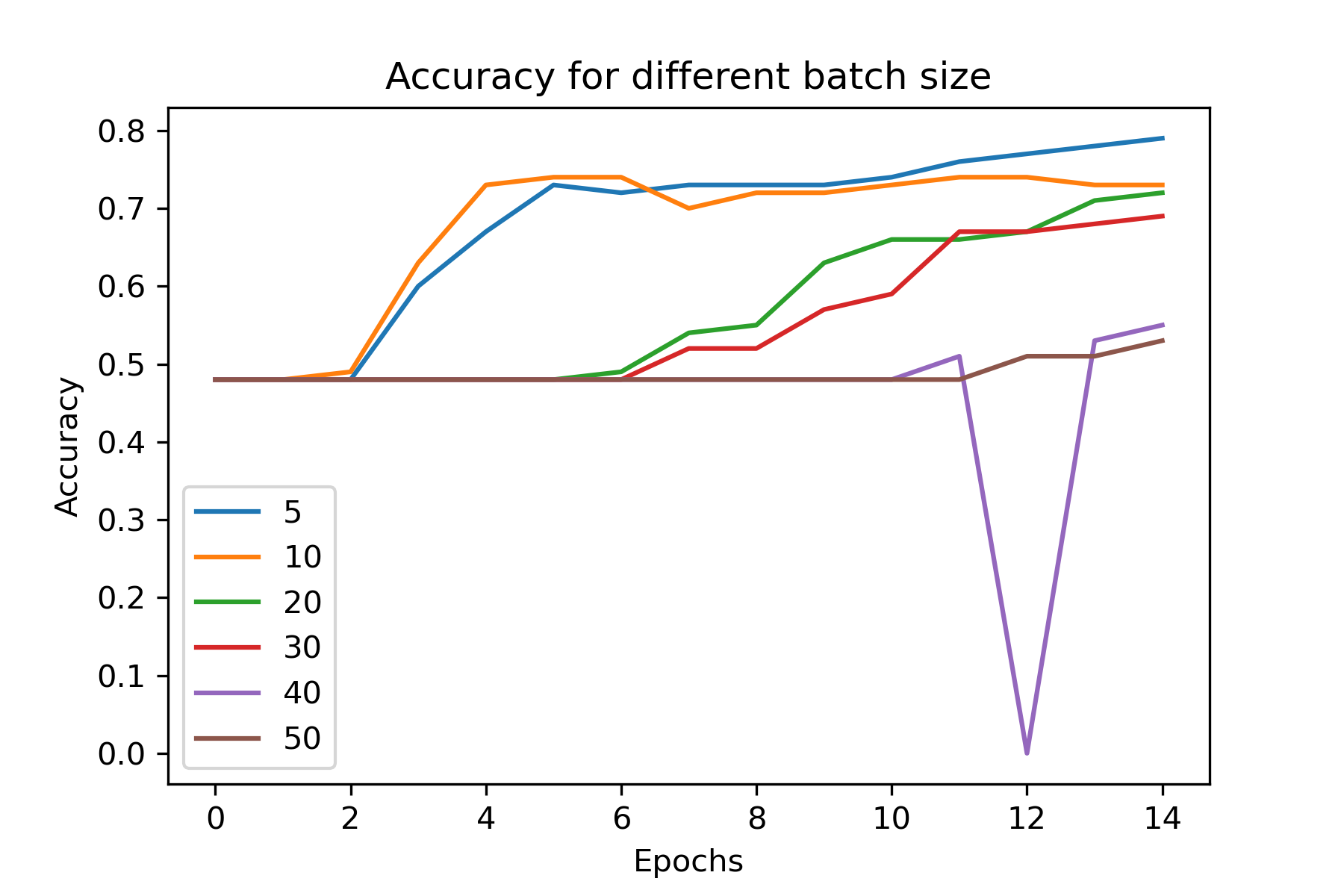}
    \caption{Accuracy for different batch sizes.}
  \end{minipage}
  \hfill
  \begin{minipage}[b]{0.45\textwidth}
    \includegraphics[width=\textwidth]{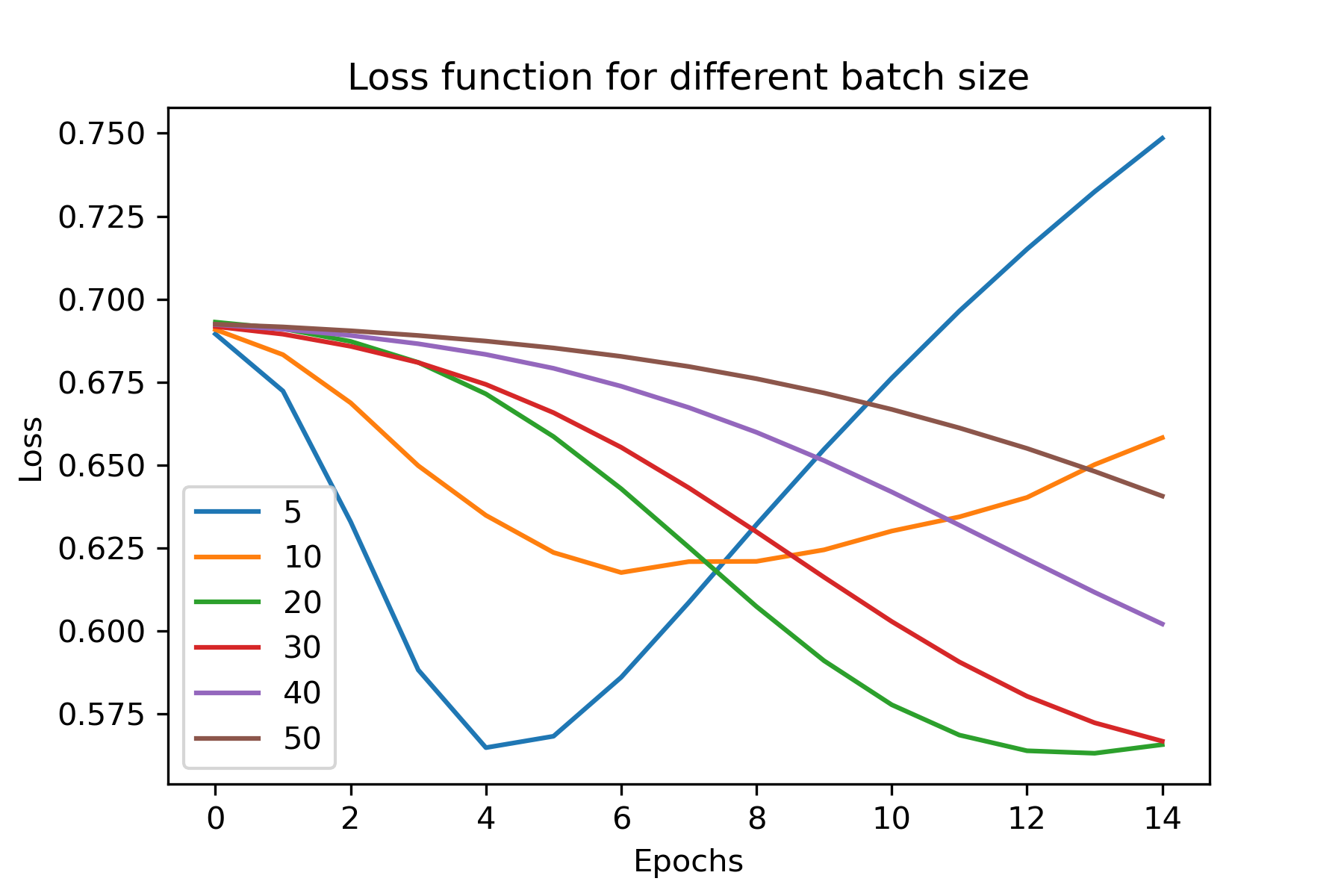}
    \caption{Loss function for different batch sizes.}
  \end{minipage}
  \label{fig:batch}
\end{figure}

Figures 6 and 7 illustrates how the batch size influences the model accuracy and loss. The loss function shows an optimum value (20) and increase for values greater or less than this optimum value. However, increasing the batch size tends to diminish the model accuracy. An alternative for taking into account both behaviours would be run a simulation for a batch size equal to 20 and consider more epochs, in order to assess if the model accuracy would be improved.

Figures 8 and 9 shows how the loss and accuracy of the model varies when considering different sequence sizes. In terms of the accuracy, this feature is improved when increasing the sequence size (i.e., considering more words in the vicinity of the interest word). Contrarily, smaller sequence lengths tend to magnify the loss. Regarding to the vocabulary size (figues 10 and 11), for all investigated values (from 1000 to 10000) the loss function is increased from the eighth epoch onward. Larger vocabularies do not exerts to much influence in the model accuracy as well. In the range investigated, the accuracy varied from 68\% to 75\%, with smaller values for the smaller vocabularies.

Figures 12 and 13 how the embedding dimension influences the machine learning modelling. The model's accuracy do not varies with different values of embedding dimension, however the loss function is significantly amplified as this parameter is increased. This behaviour suggests that higher values of embedding dimension tends to promote an overfitting. 

\begin{figure}[h!]
  \centering
  \begin{minipage}[b]{0.45\textwidth}
    \includegraphics[width=\textwidth]{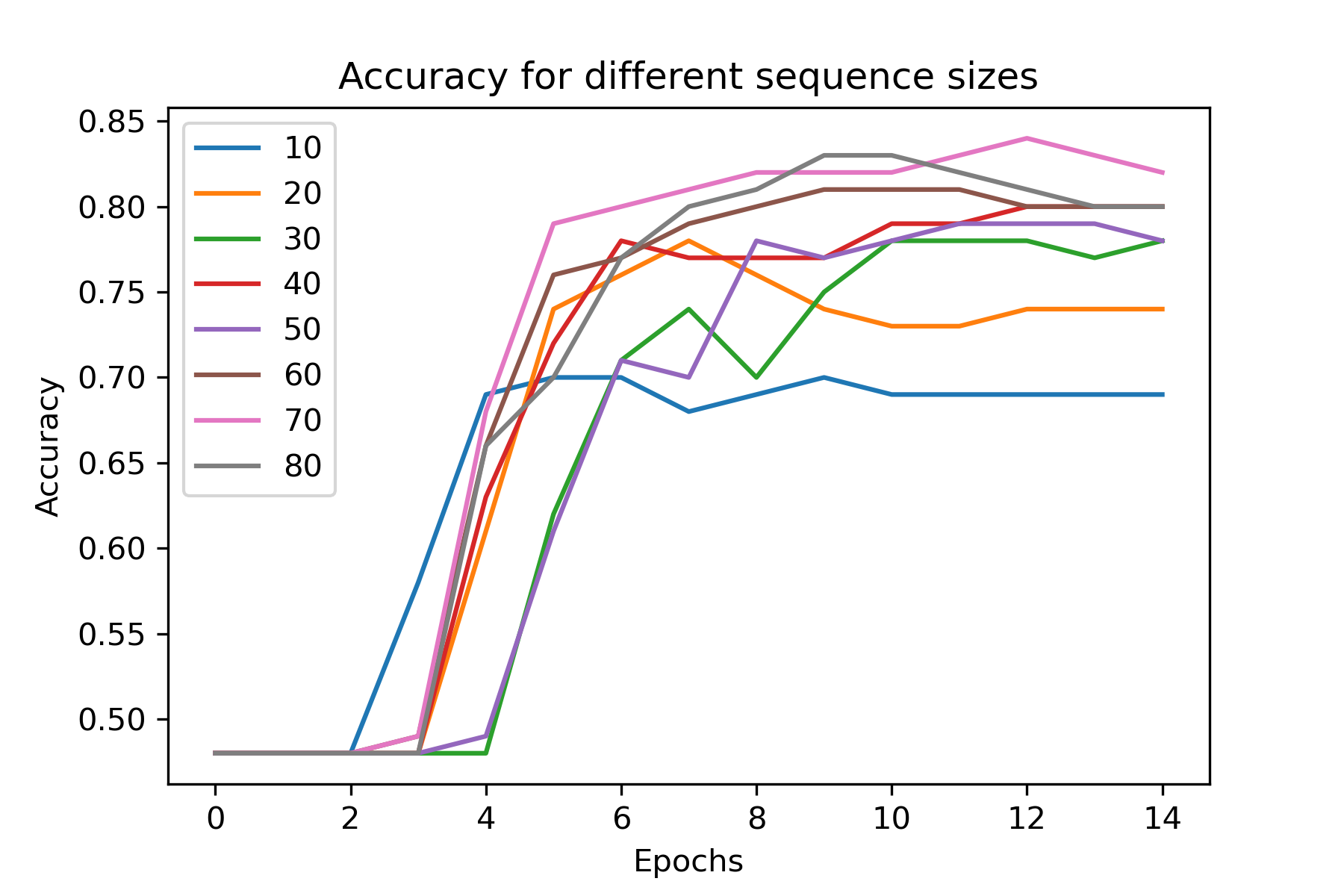}
    \caption{Accuracy for different sequence lengths.}
  \end{minipage}
  \hfill
  \begin{minipage}[b]{0.45\textwidth}
    \includegraphics[width=\textwidth]{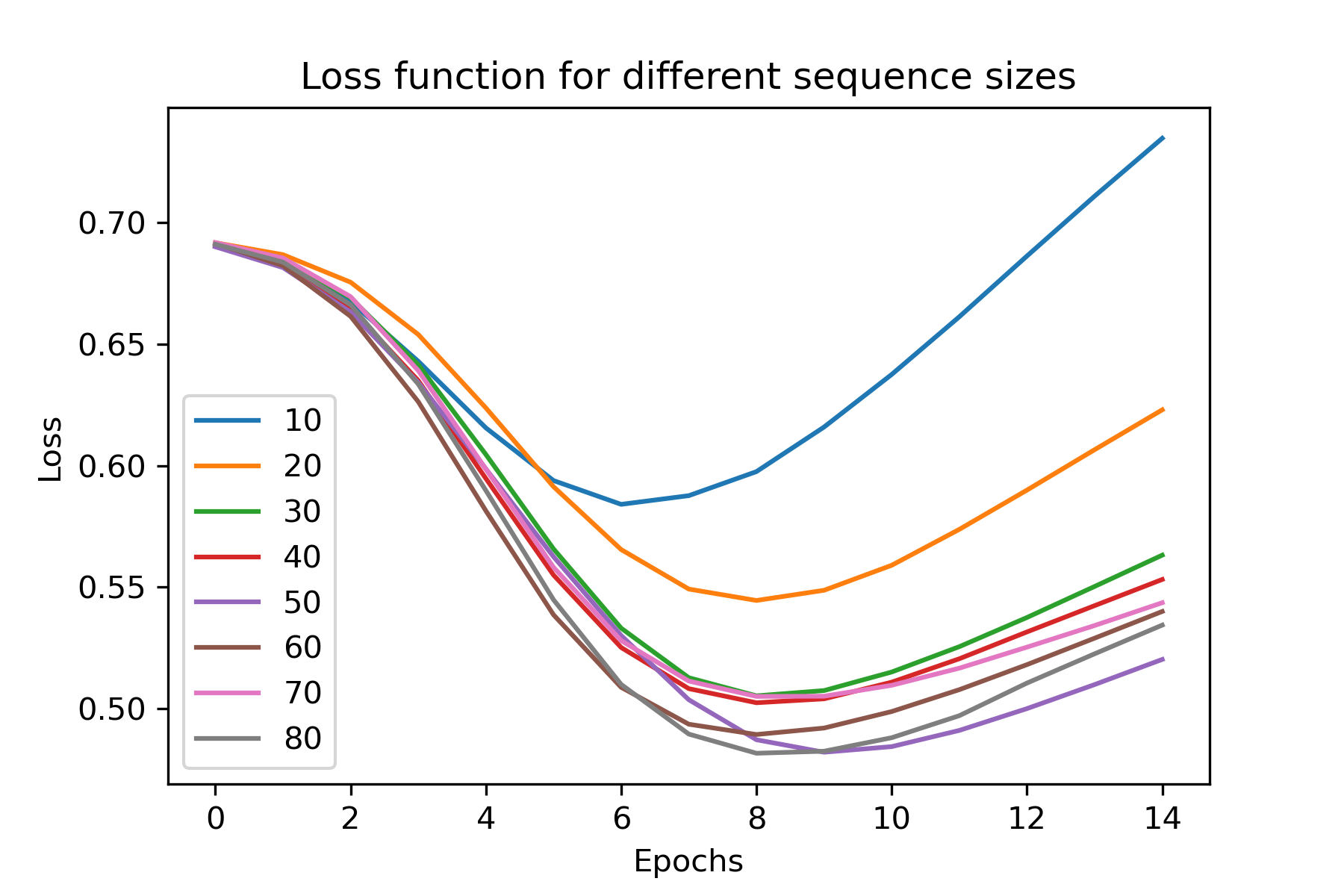}
    \caption{Loss function for different sequence lengths.}
  \end{minipage}
  \label{fig:seq}
\end{figure}

\begin{figure}[h!]
  \centering
  \begin{minipage}[b]{0.45\textwidth}
    \includegraphics[width=\textwidth]{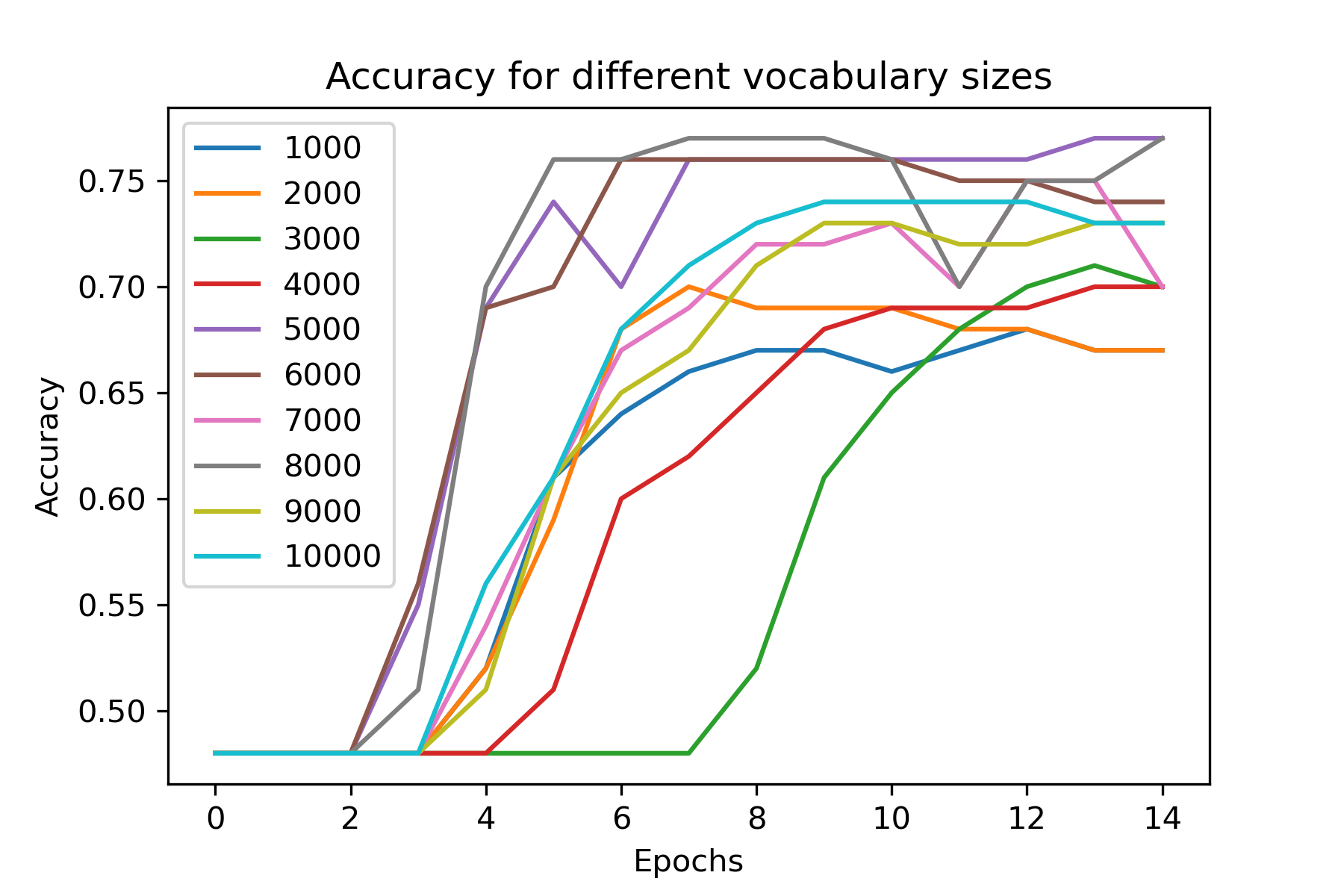}
    \caption{Accuracy for different vocabulary sizes.}
  \end{minipage}
  \hfill
  \begin{minipage}[b]{0.45\textwidth}
    \includegraphics[width=\textwidth]{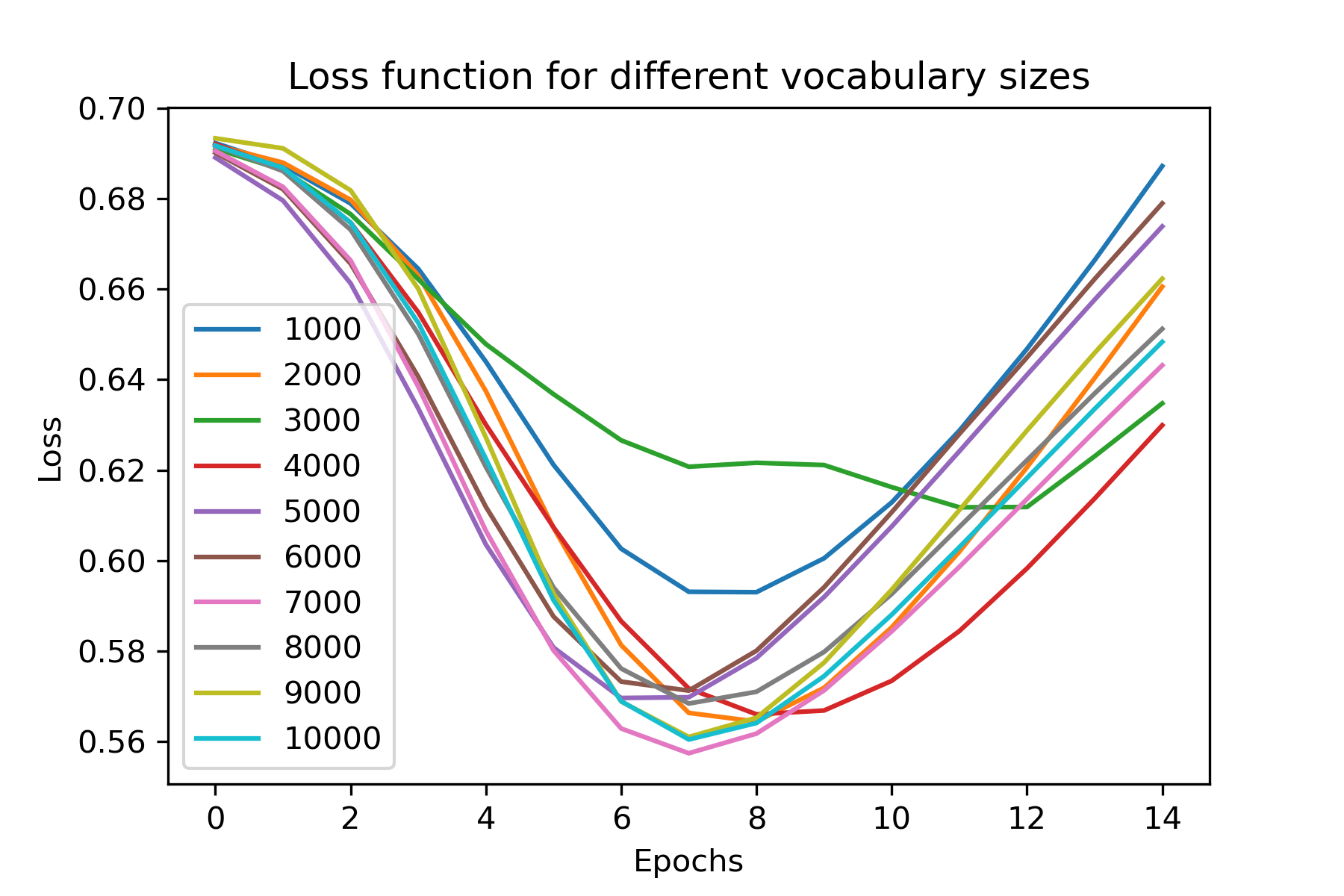}
    \caption{Loss function for different vocabulary sizes.}
  \end{minipage}
  \label{fig:vocab}
\end{figure}

\begin{figure}[h!]
  \centering
  \begin{minipage}[b]{0.45\textwidth}
    \includegraphics[width=\textwidth]{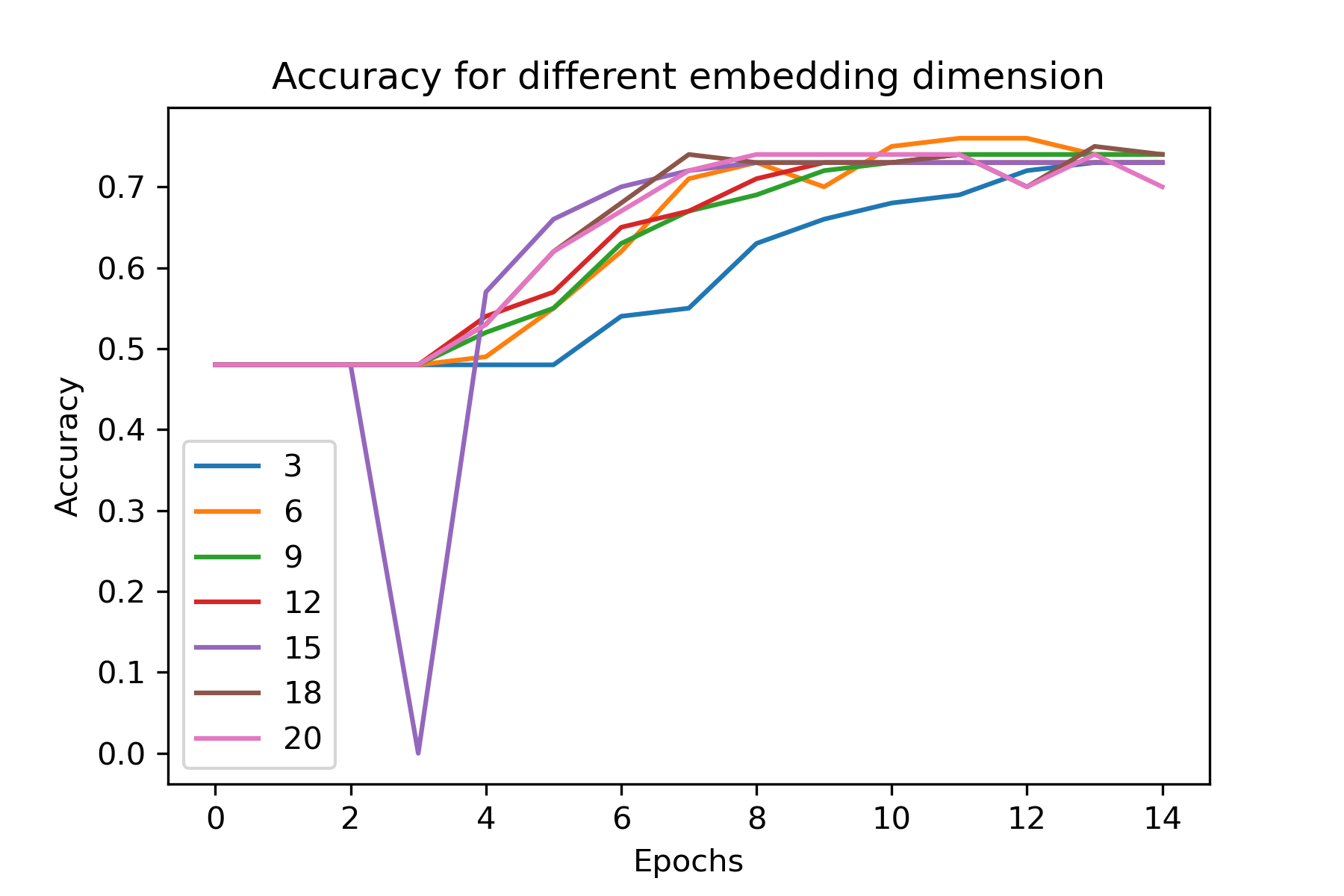}
    \caption{Accuracy for different embedding dimension.}
  \end{minipage}
  \hfill
  \begin{minipage}[b]{0.45\textwidth}
    \includegraphics[width=\textwidth]{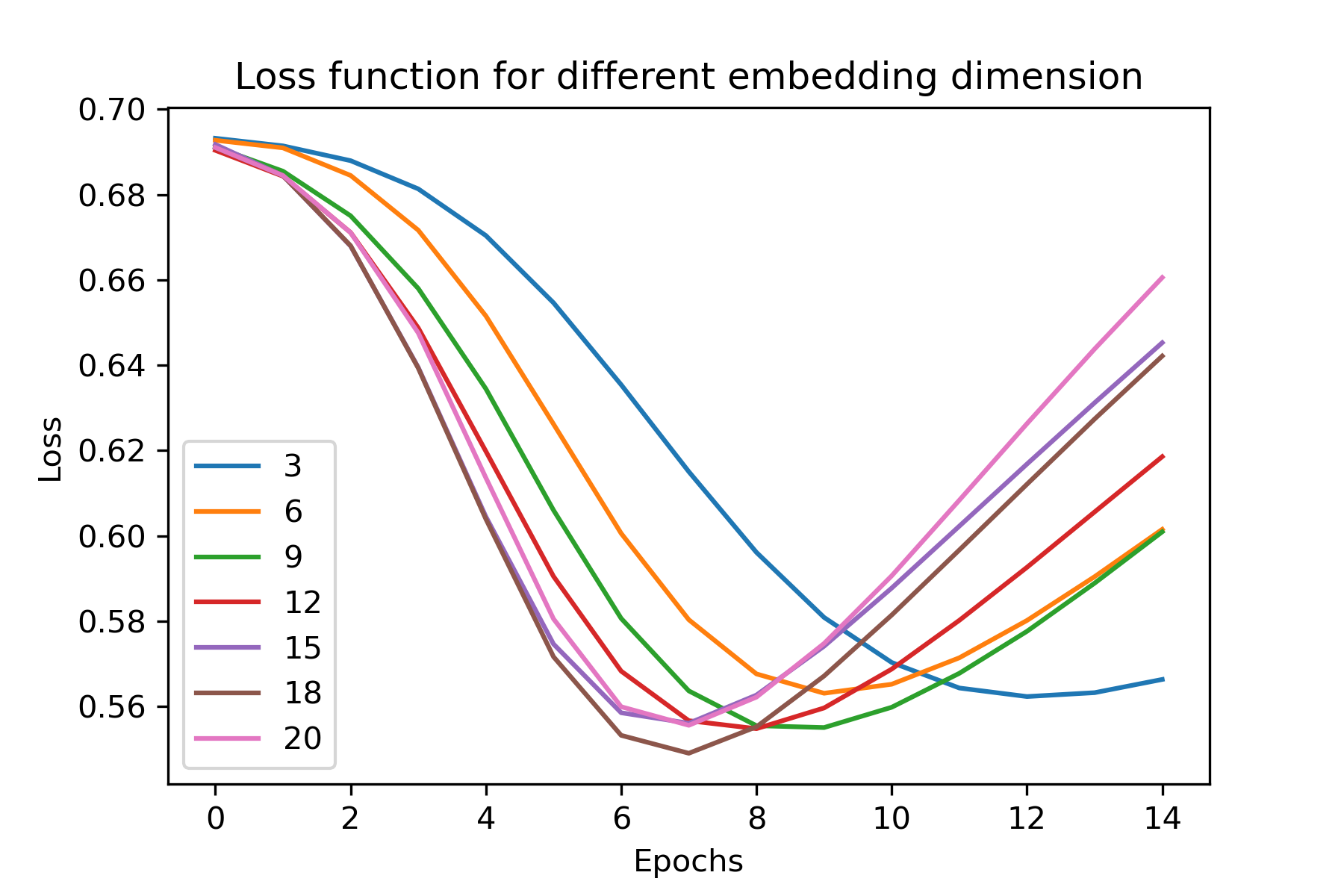}
    \caption{Loss function for different embedding dimensions.}
  \end{minipage}
  \label{fig:embed}
\end{figure}

\section{Conclusions}

This report aimed to implement machine learning approaches for natural language processing (NLP). Inspired in \textit{Sasaki et al. (2010)} and \textit{Earle et al. (2011)}, whose works explored real-time interaction on Twitter to detect natural hazards, Python routines were implemented for diverses purposes. Initially, for automatically getting content from posts in ResearchGate, and finally for using Tensorflow for computationally deal with words and make predictions/ estimates from them. To assess the validity of this mathematical (/computational) approach to natural language, Tensorboard was used to graphically visualize the nearest neighbors estimation from the Word2Vec implementation. When analysing how different parameters influences the loss and accuracy of the modeled, the batch size presented an optimum value, for which the loss is diminished. Larger sequence sizes tends to enhance the model accuracy and to diminish the loss, nevertheless from the eighth epoch onward an increasing in the loss occurs for all the investigated values, likely associated with an overfitting. Vocabularies containing from 1000 to 10000 words seemed to not exert too much influence on the mod accuracy and again from the eights epoch onward the loss is increased for all the investigated values. The embedding dimension has also shown an optimum value (3), in which the loss is less. In therms of accuracy, increasing the number of embedding dimensions seem to promote the overfitting as the loss is continuously magnified for a single value of accuracy for all values.           

\vspace*{0.5cm}

\textbf{\Large{References}}

\vspace*{0.5cm}

\noindent Sakaki, T., Okazaki, M., \& Matsuo, Y. 2010. Earthquake Shakes Twitter Users: Real-Time Event Detection by Social Sensors. \textit{Proceedings of the 19th International Conference on World Wide Web}, WWW ’10, 851–60. 

\noindent Earle, P. S., Bowden, D. C., \& Guy, M. 2011. “Twitter Earthquake Detection: Earthquake Monitoring in a Social World.” Annals of Geophysics 54 (6): 708–15.

\noindent Word Embeddings, TensorFlow Tutorials, \url{https://www.tensorflow.org/tutorials/text/word_embeddings} 

\noindent Word2Vec, TensorFlow Tutorials, \url{https://www.tensorflow.org/tutorials/text/word2vec}

\end{document}